\documentclass[letterpaper]{article} % DO NOT CHANGE THIS
\usepackage[preprint]{aaai2027}  % Use AAAI formatting without an AAAI publication claim.
\usepackage[hyphens]{url}  % DO NOT CHANGE THIS
\usepackage{graphicx} % DO NOT CHANGE THIS
\usepackage{natbib}  % DO NOT CHANGE THIS AND DO NOT ADD ANY OPTIONS TO IT
\usepackage{caption} % DO NOT CHANGE THIS AND DO NOT ADD ANY OPTIONS TO IT
\usepackage{algorithm}
\usepackage{newfloat}
\usepackage{listings}
\DeclareCaptionStyle{ruled}{labelfont=normalfont,labelsep=colon,strut=off} % DO NOT CHANGE THIS
\floatstyle{ruled}
\newfloat{listing}{tb}{lst}{}
\floatname{listing}{Listing}

\usepackage{booktabs}
\usepackage[table]{xcolor}
\usepackage[most]{tcolorbox}
\usepackage{enumitem}

\definecolor{tablegray}{RGB}{100,100,100}

\usepackage{subcaption}

\usepackage{amsmath}

\usepackage{amsmath,amssymb,bm}
\usepackage{algorithm}
\usepackage{algpseudocode}

\usepackage{pifont}
\usepackage{multirow}

\newcommand{\cmark}{\ding{51}}
\newcommand{\xmark}{\ding{55}}

\newtcolorbox{promptbox}[1]{
  enhanced,
  breakable,
  colback=white,
  colframe=black!35,
  colbacktitle=black!8,
  coltitle=black,
  title={#1},
  title after break={#1 (continued)},
  fonttitle=\small\bfseries,
  fontupper=\small,
  fontlower=\small,
  toptitle=1.2mm,
  bottomtitle=1.2mm,
  lefttitle=2.2mm,
  righttitle=2.2mm,
  boxrule=0.45pt,
  arc=0.8mm,
  outer arc=0.8mm,
  left=2.2mm,
  right=2.2mm,
  top=1.8mm,
  bottom=1.8mm,
  boxsep=0pt,
  width=\linewidth,
  before skip=8pt,
  after skip=8pt,
  segmentation style={solid,black!20,line width=0.4pt},
  before upper={\setlength{\parindent}{0pt}\setlength{\parskip}{3pt}},
  before lower={\setlength{\parindent}{0pt}\setlength{\parskip}{3pt}}
}
\newcommand{\promptrole}[1]{\textbf{#1}\par\smallskip}

\algrenewcommand\alglinenumber[1]{\scriptsize #1:}

\title{STEGNav: Spatio-Temporal Event Graph Reasoning for Multimodal Lifelong Object Navigation}
\author{
    Yang Chen\textsuperscript{\rm 1,\rm 2}\equalcontrib,
    Zhenyu Huang\textsuperscript{\rm 1,\rm 2}\equalcontrib,
    Wenbo Fu\textsuperscript{\rm 1,\rm 2},
    Danyang Peng\textsuperscript{\rm 1,\rm 2},\\
    Shi-Yu Tian\textsuperscript{\rm 1,\rm 3},
    Kun-Yang Yu\textsuperscript{\rm 1,\rm 3},
    Lan-Zhe Guo\textsuperscript{\rm 1,\rm 2}\corresponding
}
\affiliations{
    \textsuperscript{\rm 1}State Key Laboratory of Novel Software Technology, Nanjing University\\
    \textsuperscript{\rm 2}School of Intelligence Science and Technology, Nanjing University\\
    \textsuperscript{\rm 3}School of Artificial Intelligence, Nanjing University
}

\begin{document}

\maketitle

\begin{abstract}
Multimodal lifelong navigation requires an agent to autonomously explore unseen environments while sequentially completing navigation tasks specified by object categories, language descriptions, or reference images. Existing methods primarily accomplish these tasks by constructing state-centric semantic scene graphs. By treating scene graphs as persistent repositories of semantic observations, these methods struggle to distinguish similar instances, jointly represent semantic targets and exploration frontiers, and effectively exploit navigation memory and trajectory experience. To address these limitations, we propose \textbf{S}patio-\textbf{T}emporal \textbf{E}vent \textbf{G}raph \textbf{N}avigation (\textbf{STEGNav}), a training-free framework that extends conventional scene graphs into spatio-temporal event graphs along complementary spatial and temporal axes. The spatial axis performs query-conditioned instance grounding and jointly represents semantic targets and occupancy-aware exploration frontiers characterized by reachability, path cost, and exploration utility. The temporal axis employs trajectory-aware dual-window memory to retain recent decision--trajectory events and verified cross-subtask navigation outcomes. A VLM-based navigation agent reasons over the resulting spatio-temporal event graph and selects either a target instance or an exploration frontier as its next navigation goal. STEGNav achieves $66.3\%$ SR and $39.7$ SPL on GOAT-Bench, as well as SR scores of $64.0\%$ and $69.4\%$ on HM3Dv1 and HM3Dv2, respectively. Ablation studies and error analyses validate the complementary effects of the two axes, demonstrating that event-driven spatio-temporal representations improve navigation reliability and cross-subtask experience reuse.
\end{abstract}

\begin{figure}[t]
  \centering
  \includegraphics[width=\linewidth]{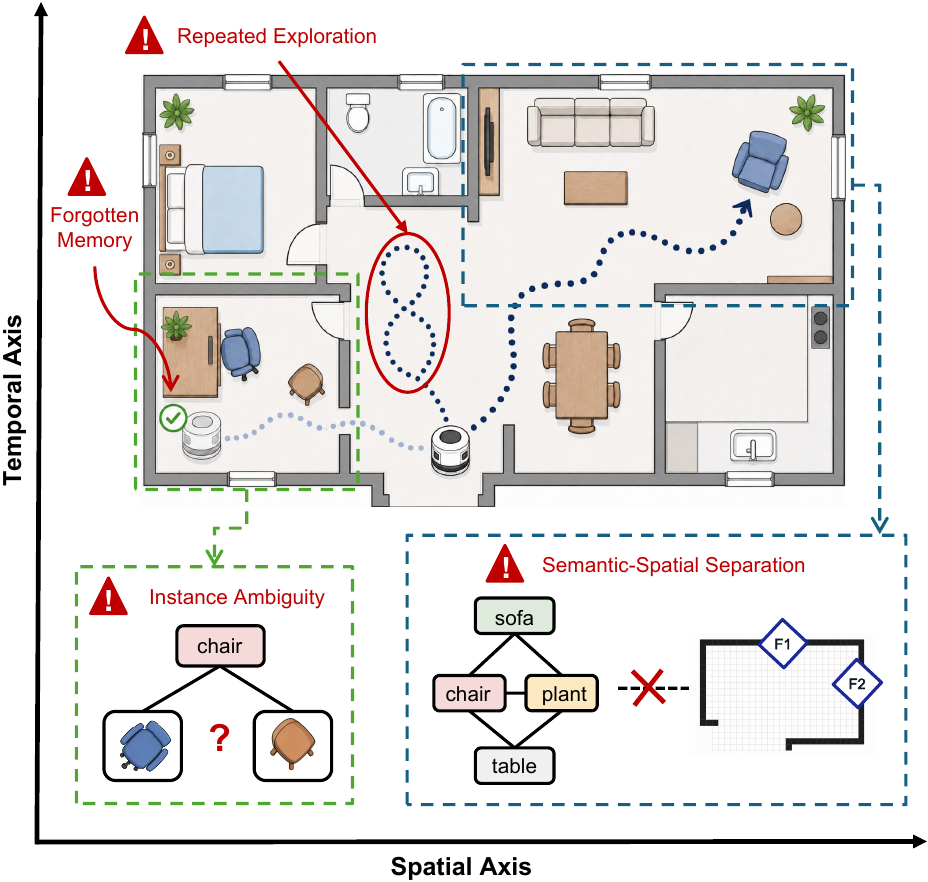}
  \caption{\textbf{Limitations of state-centric scene graphs.} Spatially, they suffer from instance ambiguity and semantic--spatial separation; temporally, the lack of explicit navigation history leads to memory loss and repeated exploration.}
  \label{fig:intro}
\end{figure}

\section{Introduction}
\label{sec:introduction}

Embodied navigation requires an agent to understand a target, explore an unseen environment, and reach the target location~\cite{habitat,objectnav}. Recent advances in Vision Language Models (VLMs) enable zero-shot navigation toward open-vocabulary targets without task-specific training~\cite{vlfm,zson,l3mvn,pixnav,spnet,compassnav,sgm}. Meanwhile, navigation tasks are evolving from isolated single-target search to multimodal lifelong navigation, where an agent must sequentially complete multiple subtasks specified by object categories, language descriptions, or reference images~\cite{goat,onemap}. This setting requires the agent not only to interpret queries from different modalities, but also to continually explore the environment and reuse knowledge accumulated across subtasks~\cite{ssmg,astranav}.

Scene graphs provide a natural representation for lifelong navigation by organizing observed objects and their semantic relations into a graph structure~\cite{conceptgraphs,scenegraphfusion,sgnav}. Recent methods further incorporate multimodal observations and persistent memory into scene graphs to support open-vocabulary reasoning and long-horizon navigation~\cite{evomem,msgnav}. However, existing scene graphs are primarily \emph{state-centric} and exhibit three major limitations: (1) they describe the objects currently present in the environment but do not explicitly distinguish among individual instances; (2) they fail to represent how the agent reached its current state or the potential spatial information associated with target states; and (3) they cannot effectively use prior decision experience to improve subsequent navigation. In contrast, multimodal lifelong navigation is inherently \emph{event-driven}, as each navigation decision changes both the agent's spatial knowledge and the historical experience available for future navigation.

To address these limitations, we propose \textbf{S}patio-\textbf{T}emporal \textbf{E}vent \textbf{G}raph \textbf{N}avigation (\textbf{STEGNav}), a training-free framework for multimodal lifelong navigation. STEGNav extends a conventional scene graph along complementary spatial and temporal axes, transforming it into a \emph{spatio-temporal event graph}. As illustrated in Fig.~\ref{fig:intro}, the spatial axis constructs a query-conditioned, multi-layer semantic--navigational graph. Query-conditioned instance grounding associates the current query with individual scene instances, enabling the agent to distinguish different objects within the same category. Building upon this grounding, occupancy-aware frontier grounding integrates semantic objects, exploration frontiers, reachability, path costs, and exploration utility into a unified graph structure. The navigation VLM can therefore directly select either a target-instance node or an exploration-frontier node from the same structured representation.

The temporal axis introduces trajectory-aware dual-window memory to organize navigation experience at two timescales. The short-term window records recent high-level decisions, executed trajectories, traversed regions, and visited frontiers, allowing the VLM to identify repeated navigation patterns. The long-term window retains only verified outcomes from previous subtasks and reconnects them to the evolving scene graph. By combining the two windows, the navigation policy can jointly leverage query-relevant instances, the current navigation state, recent trajectories, and cross-subtask experience within a unified representation.

We evaluate STEGNav on GOAT-Bench for multimodal lifelong navigation and on HM3D for category-level ObjectNav generalization. On the GOAT-Bench Val-Unseen split, STEGNav achieves an SR of $66.3\%$ and an SPL of $39.7$, improving SR by $3.9$ percentage points over the strongest competing method. On HM3Dv1 and HM3Dv2, STEGNav achieves SR of $64.0\%$ and $69.4\%$, respectively. Error analysis further shows that STEGNav reduces the total number of failures by $34.1\%$, including reductions of $53.8\%$ in instance confusion and $40.0\%$ in inefficient exploration.

Our main contributions are summarized as follows:
\begin{itemize}
    \item We propose STEGNav, a training-free multimodal lifelong navigation framework that transforms a state-centric scene graph into an event-driven spatio-temporal representation.

    \item We develop a spatial axis that unifies query-conditioned instance grounding with occupancy-aware semantic--frontier relations, together with a temporal axis that organizes recent navigation trajectories and verified cross-subtask outcomes through dual-window memory.

    \item We conduct extensive experiments on GOAT-Bench and HM3D. The results demonstrate that STEGNav improves the reliability of multimodal navigation and generalizes effectively to ObjectNav, while ablation studies and error analysis further validate the effectiveness of the proposed spatial and temporal components.
\end{itemize}

\section{Related Work}

\subsection{Object Navigation}

Object navigation requires an embodied agent to locate a target in an unseen environment and execute navigation actions toward it. Early methods primarily trained navigation policies through reinforcement learning or imitation learning~\cite{ddppo,aux,habitatweb,semexp}. Although these methods achieve strong performance within their training distributions, they typically rely on large-scale supervision and exhibit limited generalization to unseen environments and open-vocabulary targets. More recently, zero-shot navigation methods have leveraged pretrained vision-language models~\cite{clip,yoloworld} to establish semantic correspondences between open-vocabulary targets and visual observations. CoW employs CLIP-based visual matching to identify target-relevant regions, whereas VLFM constructs a target-conditioned value map to guide frontier-based exploration~\cite{cow,vlfm}.

Zero-shot object navigation has further evolved from isolated single-target search~\cite{objectnav,hm3dsem,hm3dovon,apexnav,imagegoal,ieve,consistnav, aerrnav,vlmnav,trajrag,glmap} to long-horizon navigation involving multimodal target sequences~\cite{goat,multion,tango,dynavlm}. Under this setting, an agent must not only interpret targets specified through different modalities without task-specific training, but also continually retain and reuse environmental knowledge acquired across multiple subtasks.

\subsection{Scene-Graph-Based Lifelong Navigation}

Scene graphs provide structured environmental representations by organizing objects, regions, and their semantic and spatial relations. Compared with dense geometric maps, scene graphs explicitly encode semantic knowledge, thereby providing more direct support for high-level reasoning and long-horizon planning~\cite{ssmg,evomem,dsg,sgnav,unigoal,saynav,3dmem,hsan}. MTU3D combines scene graphs with persistent memory to support lifelong multi-target navigation~\cite{mtu3d}. Building upon this paradigm, MSGNav retains visual evidence in a multimodal 3D scene graph to support open-vocabulary reasoning and final viewpoint selection~\cite{msgnav}.

Existing methods treat scene graphs as persistent, state-centric repositories, limiting their ability to distinguish similar instances, integrate semantic targets with exploration frontiers, and exploit trajectories and cross-subtask experience. STEGNav instead extends scene graphs along complementary spatial and temporal axes. The spatial axis associates queries with candidate instances and integrates navigation status and exploration utility, while the temporal axis maintains recent trajectories and retrieves query-relevant outcomes. Together, they form a reusable, event-driven spatio-temporal representation for multimodal lifelong navigation.

\begin{figure*}[t!]
  \centering
  \includegraphics[width=\textwidth]{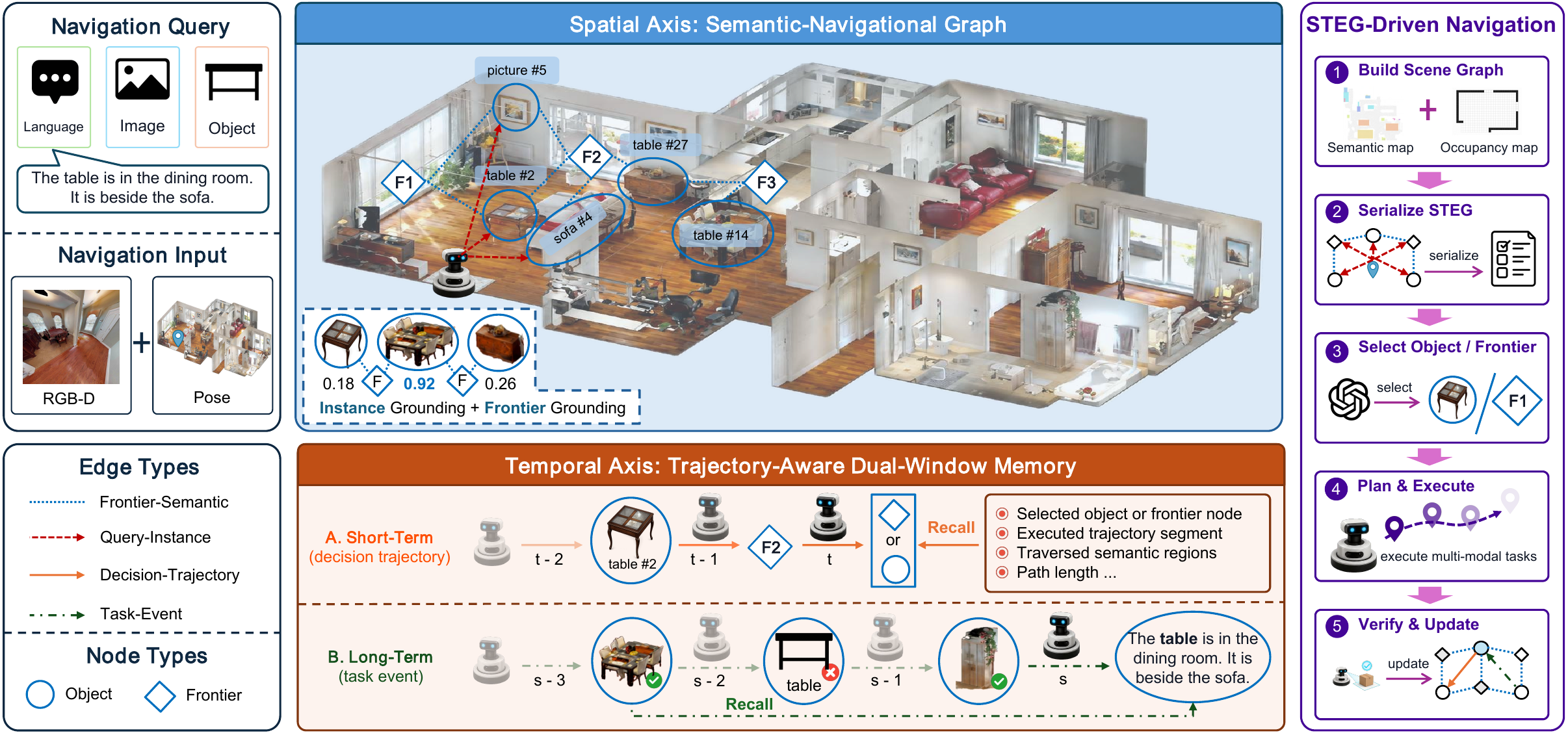}
  \caption{\textbf{Overview of STEGNav.} The spatial axis unifies query-conditioned instances with occupancy-aware frontiers, while the temporal axis organizes recent trajectories and verified cross-subtask outcomes. Their integration forms a spatio-temporal event graph for VLM-based navigation, verification, and continual update.}
  \label{fig:framework}
\end{figure*}

\section{Preliminaries}
\label{sec:preliminaries}

\subsection{Scene-Graph-Based Multimodal Lifelong Navigation}
In an unseen scene $\mathcal{S}$, an agent sequentially completes a set of multimodal navigation subtasks
$\{q_k\}_{k=1}^{K}$ within a single episode. Each query $q_k=(c_k,m_k)$ specifies a target category $c_k$ and a query modality $m_k\in\{\texttt{object},\texttt{language},\texttt{image}\}$. The \texttt{object} modality specifies a category-level target, whereas the language and image modalities refer to a particular physical instance. At each time step, the agent receives an RGB-D observation and outputs a navigation action. A subtask is considered successful if the agent issues \texttt{stop} within $1.0\,\mathrm{m}$ of an instance satisfying the current query.

The agent maintains an online semantic scene graph $\mathcal{G}_t=(\mathcal{V}_t,\mathcal{E}_t)$. Each object node is represented as
\begin{equation}
    v_i^t=
    \left(
    i,c_i,B_i^t,\ell_i^t,\bm f_i^t,I_i^t
    \right),
\end{equation}
where $i$ is a persistent identifier, $c_i$ is the semantic category, $B_i^t$ is the current 3D bounding box, $\ell_i^t$ is the room label, $\bm f_i^t$ is the aggregated appearance feature, and $I_i^t$ is a representative RGB crop of the instance. Matched observations update existing nodes, whereas unmatched observations instantiate new  ones. The current center of node $i$ is given by $\bm p_i^t=\operatorname{center}(B_i^t)$.

The agent additionally maintains an occupancy map $\mathbf{O}_t^{\mathrm{occ}}$. At each step, the depth observation is projected into a 3D point cloud, transformed into the global coordinate frame using the agent pose, and accumulated into a 2D grid that distinguishes free, occupied, and unexplored regions. Frontier cells are identified at the boundaries between explored free space and unexplored space, and adjacent frontier cells are clustered into frontier regions to form $\mathcal{F}_t$, with their exploration gains estimated from the surrounding unexplored area.

At each high-level decision step, a VLM-based category prefilter $P$ retains the scene nodes whose semantic categories are relevant to the current target. The retained subgraph is serialized, while the exploration frontiers are provided as a separate input:
\begin{equation}
\begin{aligned}
    \mathcal{V}_{k,t}^{\mathrm{pre}}
    &=P(c_k,\mathcal{G}_t),\\
    y_{k,t}
    &=\pi_{\mathrm{VLM}}\!\left(
      q_k,
      \operatorname{Ser}\!\left(
      \mathcal{G}_t[
      \mathcal{V}_{k,t}^{\mathrm{pre}}]
      \right),
      \mathcal{F}_t
      \right).
\end{aligned}
\label{eq:baseline_policy}
\end{equation}
Although $P$ preserves nodes related to the target category, it cannot distinguish different physical instances belonging to the same category. Moreover, the scene graph and the occupancy-derived frontiers are represented independently, preventing the navigation policy from jointly reasoning about target identity, traversability, and exploration utility.

\subsection{Motivation}
Existing scene-graph-based navigation policies typically preserve objects related to the target category while representing occupancy-derived frontiers as separate information. This design introduces two limitations. Spatially, category-level filtering cannot distinguish among same-category instances, while the separation between semantic nodes and exploration frontiers prevents joint reasoning over target identity, traversability, and exploration utility. Temporally, the online scene graph records only the current environmental state without explicitly preserving the navigation process, which can lead to repeated exploration and the loss of previously verified targets. These limitations motivate a spatio-temporal event graph that unifies semantic instances with occupancy-aware spatial information along the spatial axis and organizes recent navigation trajectories and verified historical outcomes along the temporal axis.

\section{Method}
\label{sec:method}

\subsection{Method Overview}
\label{sec:method_overview}

For the current subtask $q_k$, we first construct a query event
$u_k=(q_k,c_k,m_k)$. The spatial axis builds a query-conditioned semantic--navigational graph
$\mathcal{G}_{k,t}^{\mathrm{spa}}$ that jointly represents semantic instances, their navigation relations, and exploration gains. The temporal axis maintains recently executed decisions and their trajectories in a short-term window
$\mathcal{M}_{k,t}^{\mathrm{st}}$, while retaining verified navigation outcomes from previous subtasks in a long-term window
$\mathcal{M}_{k}^{\mathrm{lt}}$.

Together, these components form the spatio-temporal event graph
\begin{equation}
\begin{aligned}
    \mathcal{V}_{k,t}^{\mathcal H}
    &=
    \mathcal{V}_{k,t}^{\mathrm{spa}}
    \cup\mathcal{M}_{k,t}^{\mathrm{st}}
    \cup\mathcal{M}_{k}^{\mathrm{lt}},\\
    \mathcal{E}_{k,t}^{\mathcal H}
    &=
    \mathcal{E}_{k,t}^{\mathrm{spa}}
    \cup\mathcal{E}_{k,t}^{\mathrm{st}}
    \cup\mathcal{E}_{k,t}^{\mathrm{lt}},\\
    \mathcal{H}_{k,t}
    &=
    \left(
    \mathcal{V}_{k,t}^{\mathcal H},
    \mathcal{E}_{k,t}^{\mathcal H}
    \right).
\end{aligned}
\label{eq:spatiotemporal_event_graph}
\end{equation}
The resulting graph provides a unified representation of target semantics, traversability, exploration utility, recent navigation trajectories, and verified navigation history. The framework is illustrated in Fig.~\ref{fig:framework}.

\subsection{Spatial Axis: Multi-Layer Semantic--Navigational Graph}
\label{sec:spatial_axis}

\paragraph{Query-Conditioned Instance Grounding.}
We first apply the category prefilter $P$ to preserve all scene nodes semantically related to the target category. A VLM-based instance grounding module
$\Phi_{\mathrm{inst}}$ then evaluates each retained node using the current query, its visual representation, and its spatial context:
\begin{equation}
    \left(
    r_k^t(i),d_k^t(i)
    \right)
    =
    \Phi_{\mathrm{inst}}
    \left(q_k,v_i^t\right),
    \qquad
    v_i^t\in\mathcal{V}_{k,t}^{\mathrm{pre}},
    \label{eq:instance_grounding}
\end{equation}
where $r_k^t(i)$ denotes the query--instance relevance and
$d_k^t(i)$ denotes a query-conditioned language description of instance $i$.

For an image query, the grounding module evaluates the visual correspondence between the reference image and the stored appearance $I_i^t$ of each instance. For a language query, it grounds the target attributes to the appearance, room, and spatial context of each instance. For a category-level object query, no instance-level pruning is performed. For image and language queries, we retain the top-$N$ most relevant instances and construct query--instance edges carrying
\begin{equation}
    \left(
    r_k^t(i),d_k^t(i),I_i^t
    \right)
\end{equation}
as their attributes. The complete object nodes remain persistently stored in
$\mathcal{G}_t$, while only compact query-relevant information is exposed to the navigation VLM.

\paragraph{Occupancy-Aware Frontier Grounding.}
We augment the spatial representation of semantic instances with navigation information derived from the occupancy map. Adjacent frontier cells are first grouped into frontier regions, with each region represented by a frontier node
$g_r^t$. Each frontier node stores its spatial center, reachable path distance, estimated exploration gain, and recent visitation state.

Each reachable frontier is connected to nearby semantic nodes located within the same traversable region. The resulting frontier--semantic edges describe the local semantic context into which the frontier leads. The occupancy map further establishes navigation edges from the current agent node to both object and frontier nodes. Each navigation edge records the reachability and geodesic path cost of the corresponding target. Consequently, a semantic object node represents not only category- and instance-level information, but also its current navigation status.

Let $\mathcal{V}_{k,t}^{\mathrm{sem}}$ contain the currently grounded target instances, the instances recalled along the temporal axis, and the compact semantic anchors required to contextualize the exploration frontiers. The resulting spatial graph is defined as
\begin{equation}
\begin{aligned}
    \mathcal{G}_{k,t}^{\mathrm{spa}}
    &=
    \left(
    \mathcal{V}_{k,t}^{\mathrm{spa}},
    \mathcal{E}_{k,t}^{\mathrm{spa}}
    \right),\\
    \mathcal{V}_{k,t}^{\mathrm{spa}}
    &=
    \{u_k,v_t^{\mathrm{agent}}\}
    \cup\mathcal{V}_{k,t}^{\mathrm{sem}}
    \cup\mathcal{V}_{t}^{\mathrm{fr}},\\
    \mathcal{E}_{k,t}^{\mathrm{spa}}
    &=
    \mathcal{E}_{k,t}^{\mathrm{sem}}
    \cup\mathcal{E}_{k,t}^{\mathrm{qry}}
    \cup\mathcal{E}_{k,t}^{\mathrm{nav}}.
\end{aligned}
\label{eq:multilayer_spatial_graph}
\end{equation}
Here, $\mathcal{E}_{k,t}^{\mathrm{sem}}$ contains the semantic and spatial relations inherited from the scene graph,
$\mathcal{E}_{k,t}^{\mathrm{qry}}$ connects the query event to the grounded target instances, and
$\mathcal{E}_{k,t}^{\mathrm{nav}}$ represents occupancy-derived reachability, path costs, and frontier connections. Candidate objects and exploration frontiers are thus represented within the same graph structure rather than being provided to the VLM as separate inputs.

\subsection{Temporal Axis: Trajectory-Aware Dual-Window Memory}
\label{sec:temporal_axis}

The temporal axis jointly maintains the recent navigation process within the current subtask and navigation outcomes retained across subtasks.

\paragraph{Short-Term Decision-Trajectory Window.}
After each high-level decision is executed, we append a short-term event containing the selected object or frontier node, the executed trajectory segment, the traversed semantic regions, the path length, and the execution outcome. Object-directed decisions reference persistent scene-node identifiers, whereas frontier-directed decisions retain their corresponding semantic anchors and spatial regions.

Rather than storing the complete sequence of low-level poses, each trajectory segment is compactly represented by its start and end states, traversed graph nodes, and occupancy coverage. Consecutive events are connected by temporal edges, and each event is linked to the object or frontier node that triggered the corresponding trajectory. The stored trajectories are subsequently projected onto the current spatial graph to update frontier visitation states and identify previously traversed regions. This representation explicitly exposes repeated paths and explored frontiers to the VLM without imposing a hard exploration constraint.

The short-term window retains the latest
$W_{\mathrm{st}}$
executed high-level decisions together with their corresponding trajectories. Repeated decisions involving the same object or spatial region remain as distinct events, thereby preserving the actual navigation sequence. The short-term window is cleared when the current subtask terminates.

\paragraph{Long-Term Task-Event Window.}
The long-term window stores outcomes that receive positive VLM verification. After navigating toward instance $i_k^*$, the outcome is verified by
\begin{equation}
    w_k
    =
    \mathbf{1}\!\left[
    i_k^{\mathrm{term}}=i_k^*
    \land o_k=+
    \land d_k\leq\delta
    \right],
    \label{eq:positive_verification}
\end{equation}
where $i_k^{\mathrm{term}}$ denotes the instance associated with the terminal observation, $o_k\in\{+,-\}$ is the VLM judgment based on the query and terminal visual evidence, and $d_k$ is the distance from the terminal agent pose to the selected node. No simulator ground truth is used during verification.

When $w_k=1$, the outcome is stored as an event anchor $a_k$, containing the reached instance identifier, target category, query modality, historical query, terminal appearance feature, and room label, without fixed coordinates. We set $W_{\mathrm{lt}}=8$. These anchors serve only as recall cues and are re-associated with the current scene graph rather than treated as ground-truth targets.

Short-term object events are reconnected through persistent scene-node identifiers. For a long-term anchor $a$, an exact identifier match is accepted directly. If the identifier is unavailable because of graph merging or node re-instantiation, we compute
\begin{equation}
    \rho_t(i,a)
    =
    \left\langle
    \bar{\bm f}_i^t,
    \bar{\bm f}_a
    \right\rangle.
    \label{eq:event_node_score}
\end{equation}
The association is accepted when $\rho_t(i,a)\geq\gamma_{\mathrm{hi}}$ and the available room labels are consistent, and rejected when $\rho_t(i,a)\leq\gamma_{\mathrm{lo}}$. Ambiguous cases are resolved by a lightweight VLM verifier.

% Long-term events must additionally satisfy category consistency and modality compatibility with the current query. A category-level historical event can support a subsequent category query, but it cannot provide instance-specific evidence for a language- or image-conditioned query.

\subsection{Spatio-Temporal Event-Graph Navigation}
\label{sec:event_graph_navigation}

At each decision step, the currently grounded instances are combined with the scene nodes referenced by the two temporal windows. Temporally recalled nodes are re-grounded using the latest scene graph and occupancy map, ensuring that their appearance, position, and traversability reflect the current state. Short-term trajectory events are then connected to the current semantic and frontier nodes through persistent identifiers, semantic anchors, and spatial overlap, producing the current trajectory context $\mathcal{Z}_t$.

After these connections are established, the updated event graph is serialized as
\begin{equation}
\begin{aligned}
    \bm h_{k,t}^{\mathrm{evt}}
    &=
    \operatorname{Ser}\!\left(
    \widetilde{\mathcal{H}}_{k,t}
    \right),\\
    y_{k,t}
    &=
    \pi_{\mathrm{VLM}}\!\left(
    q_k,\bm h_{k,t}^{\mathrm{evt}}
    \right).
\end{aligned}
\label{eq:event_graph_policy}
\end{equation}
Unlike the baseline in Eq.~\eqref{eq:baseline_policy}, exploration frontiers are directly incorporated into
$\widetilde{\mathcal{H}}_{k,t}$ rather than being provided as an independent list. The serialized graph exposes query-conditioned instance descriptions, relevance scores, current reachability, frontier exploration utility, recent decision trajectories, and verified historical outcomes to the VLM.

The navigation VLM directly selects either a target-instance node or an exploration-frontier node. When an object node is selected, the agent navigates toward its current center. When a frontier node is selected, the agent follows the corresponding occupancy-derived path. After either decision, the executed trajectory and its outcome are appended to the short-term window. A verified object-navigation success additionally produces a long-term event anchor.
Algorithm~\ref{alg:spatiotemporal_policy} summarizes the complete navigation policy.

\begin{algorithm}[t]
\caption{Trajectory-Aware Spatio-Temporal Event-Graph Navigation}
\label{alg:spatiotemporal_policy}
\small
\begin{algorithmic}[1]
\Require Query $q_k$, scene graph $\mathcal{G}_t$,
occupancy map $\mathbf{O}_t^{\mathrm{occ}}$,
long-term memory $\mathcal{M}_{k}^{\mathrm{lt}}$
\Ensure Updated long-term memory
$\mathcal{M}_{k+1}^{\mathrm{lt}}$

\State $u_k\gets(q_k,c_k,m_k)$
\State $\mathcal{M}^{\mathrm{st}}\gets\varnothing$
\State $\mathcal{M}_{k+1}^{\mathrm{lt}}
       \gets\mathcal{M}_{k}^{\mathrm{lt}}$

\While{$q_k$ is active and the step budget remains}
    \State $\mathcal{V}^{\mathrm{pre}}\gets
       P(c_k,\mathcal{G}_t)$

    \State $(\mathcal{R}^{\mathrm{st}},
       \mathcal{R}^{\mathrm{lt}},
       \mathcal{Z}_t)\gets
       \operatorname{TemporalLink}
       (\mathcal{M}^{\mathrm{st}},
        \mathcal{M}_{k}^{\mathrm{lt}},
        \mathcal{G}_t)$

    \State $\mathcal{V}^{\mathrm{tar}}\gets
       \operatorname{InstanceGround}
       (q_k,\mathcal{V}^{\mathrm{pre}},
        \mathcal{R}^{\mathrm{st}},
        \mathcal{R}^{\mathrm{lt}})$

    \State $\mathcal{F}_t\gets
       \operatorname{Frontier}
       (\mathbf{O}_t^{\mathrm{occ}})$

    \State $\mathcal{G}^{\mathrm{spa}}\gets
       \operatorname{SpatialGraph}
       (u_k,\mathcal{V}^{\mathrm{tar}},
        \mathcal{G}_t,\mathcal{F}_t,
        \mathbf{O}_t^{\mathrm{occ}},
        \mathcal{Z}_t)$

    \State $\widetilde{\mathcal{H}}\gets
       \operatorname{EventGraph}
       (\mathcal{G}^{\mathrm{spa}},
        \mathcal{M}^{\mathrm{st}},
        \mathcal{M}_{k}^{\mathrm{lt}})$

    \State $y\gets
       \operatorname{VLMSelect}
       (q_k,\widetilde{\mathcal{H}})$

    \If{$y$ is an object node}
        \State $i_k^*\gets\operatorname{NodeID}(y)$
        \State $(o_k,d_k,i_k^{\mathrm{term}},\xi_t)
        \gets\operatorname{Navigate}
        (\bm p_{i_k^*}^t)$
        \State Update $\mathcal{G}_t$ and
        $\mathbf{O}_t^{\mathrm{occ}}$
        \State Compute $w_k$ using
        Eq.~\eqref{eq:positive_verification}
    \Else
        \State $(o_k,\xi_t)\gets
        \operatorname{Explore}(y)$
        \State Update $\mathcal{G}_t$ and
        $\mathbf{O}_t^{\mathrm{occ}}$
        \State $w_k\gets0$
    \EndIf

    \State $e_t^{\mathrm{st}}\gets
       \operatorname{DecisionEvent}
       (y,\xi_t,o_k,\mathcal{G}^{\mathrm{spa}})$

    \State $\mathcal{M}^{\mathrm{st}}\gets
       \operatorname{AppendShort}
       (\mathcal{M}^{\mathrm{st}},
        e_t^{\mathrm{st}})$

    \If{$w_k=1$}
        \State Construct the long-term event anchor $a_k$
        \State $\mathcal{M}_{k+1}^{\mathrm{lt}}
        \gets\operatorname{UpdateLong}
        (\mathcal{M}_{k}^{\mathrm{lt}},a_k)$
        \State \textbf{break}
    \EndIf
\EndWhile

\State Clear $\mathcal{M}^{\mathrm{st}}$
\end{algorithmic}
\end{algorithm}

\section{Experiments}
\label{sec:experiments}

\subsection{Experimental Setup}
\label{sec:experimental_setup}

\paragraph{Benchmarks.}
We evaluate STEGNav on two established embodied navigation benchmarks in the Habitat simulator. GOAT-Bench~\cite{goat} evaluates multimodal lifelong navigation using object, language, and image queries. Its Val-Unseen split contains 36 scenes, with 10 episodes per scene and 5--10 sequential subtasks per episode. We use GOAT-Bench as our primary benchmark because it jointly evaluates multimodal target grounding, long-horizon exploration, and cross-subtask experience reuse. The main experimental results are evaluated on the full benchmark split. The ablation study and error analysis are conducted on a fixed subset containing the first episode from each of the 36 scenes in the GOAT-Bench Val-Unseen split.

We further evaluate category-level ObjectNav generalization on HM3D. HM3Dv1 contains 20 scenes and 2,000 episodes, while HM3Dv2 contains 36 scenes and 1,000 episodes. Both versions cover six object-goal categories. Although HM3D does not contain instance-specific multimodal queries, it provides a complementary evaluation of zero-shot target discovery and navigation efficiency.

\paragraph{Evaluation Metrics.}
Following the standard evaluation protocol, we report Success Rate (SR) and Success weighted by Path Length (SPL). A subtask is considered successful if the agent issues \texttt{stop} within $1.0\,\mathrm{m}$ of an instance satisfying the current query.

\paragraph{Implementation Details.}
We use YOLOv8-World~\cite{yoloworld}, SAM~\cite{sam} for object-category recognition and CLIP for visual feature extraction and historical appearance association. We adopt ChatGPT-5.4-mini as the VLM backbone for instance grounding, navigation-target selection, and ambiguous event verification. We set the appearance-association thresholds to
$(\gamma_{\mathrm{hi}},\gamma_{\mathrm{lo}})=(0.92,0.70)$
and use $\delta=0.8\,\mathrm{m}$ for internal outcome verification, which is stricter than the benchmark success threshold. The short-term window retains the latest $W_{\mathrm{st}}=20$ decision--trajectory events, while the long-term window stores up to $W_{\mathrm{lt}}=8$ verified subtask events. Complete prompts, graph serialization formats, module configurations, and VLM inference settings are provided in the Appendix.

\begin{table}[t!]
\centering
\caption{Main results on the GOAT-Bench Val-Unseen split. MSGNav does not report SPL under our evaluation protocol.}
\label{tab:exp1-main}
\resizebox{\linewidth}{!}{%
\begin{tabular}{lccc}
\toprule
\textbf{Method} & \textbf{Training-free} & \textbf{SR $\uparrow$} & \textbf{SPL $\uparrow$} \\ \hline
SenseAct-NN Monolithic \textcolor{tablegray}{[CVPR24]}   & \xmark & 12.3 & 6.8  \\
SenseAct-NN Skill Chain \textcolor{tablegray}{[CVPR24]}  & \xmark & 29.5 & 11.3 \\
MTU3D \textcolor{tablegray}{[ICCV25]}                    & \xmark & 47.2 & 27.7 \\ \hline
Modular CLIP on Wheels \textcolor{tablegray}{[CVPR24]}   & \cmark & 16.1 & 10.4 \\
Modular GOAT \textcolor{tablegray}{[CVPR24]}             & \cmark & 24.9 & 17.2 \\
VLMNav \textcolor{tablegray}{[NeSy25]}                   & \cmark & 20.1 & 9.6  \\
DyNaVLM \textcolor{tablegray}{[arXiv25]}                 & \cmark & 25.5 & 10.2 \\
TANGO \textcolor{tablegray}{[CVPR25]}                    & \cmark & 32.1 & 16.5 \\
3D-Mem \textcolor{tablegray}{[CVPR25]}                   & \cmark & 42.6 & 22.8 \\
EvoMemNav \textcolor{tablegray}{[arXiv26]}               & \cmark & 59.6 & 38.9 \\
MSGNav \textcolor{tablegray}{[CVPR26]}         & \cmark & 62.4 & --   \\
\midrule
\rowcolor{blue!8}
\textbf{STEGNav (Ours)}            & \textbf{\cmark} & \textbf{66.3} & \textbf{39.7} \\
\bottomrule
\end{tabular}}
\end{table}

\begin{table}[t!]
\centering
\caption{Object navigation results on the HM3D benchmarks.}
\label{tab:exp2-objnav}
\resizebox{\linewidth}{!}{%
\begin{tabular}{lccccc}
\toprule
\multirow{2}{*}{\textbf{Method}} &
\multirow{2}{*}{\textbf{Training-Free}} &
\multicolumn{2}{c}{\textbf{HM3Dv1}} &
\multicolumn{2}{c}{\textbf{HM3Dv2}} \\
\cline{3-6}
& & SR $\uparrow$ & SPL $\uparrow$ & SR $\uparrow$ & SPL $\uparrow$ \\
\midrule
ZSON \textcolor{tablegray}{[NeurIPS22]}    & \xmark & 25.5 & 12.6 & -- & -- \\
PixNav \textcolor{tablegray}{[ICRA24]}     & \xmark & 37.9 & 20.5 & -- & -- \\
SPNet \textcolor{tablegray}{[RA-L23]}      & \xmark & 31.2 & 10.1 & -- & -- \\
CompassNav \textcolor{tablegray}{[ICLR26]} & \xmark & 56.6 & 27.6 & 59.6 & 26.9 \\
SGM \textcolor{tablegray}{[CVPR24]}        & \xmark & 60.2 & 30.8 & -- & -- \\
\midrule
L3MVN \textcolor{tablegray}{[IROS23]}      & \cmark & 50.4 & 23.1 & 36.3 & 15.7 \\
VLFM \textcolor{tablegray}{[ICRA24]}       & \cmark & 52.5 & 30.4 & 63.6 & 32.5 \\
SG-Nav \textcolor{tablegray}{[NeurIPS24]}  & \cmark & 54.0 & 24.9 & 49.6 & 25.5 \\
UniGoal \textcolor{tablegray}{[CVPR25]}      & \cmark & 54.5 & 25.1 & -- & -- \\
EvoMemNav \textcolor{tablegray}{[arXiv26]} & \cmark & 59.2 & \textbf{33.6} & 63.8 & \textbf{39.4} \\
\midrule
\rowcolor{blue!8}
\textbf{STEGNav (Ours)} & \cmark & \textbf{64.0} & 29.8 & \textbf{69.4} & 28.2 \\
\bottomrule
\end{tabular}}
\end{table}

\subsection{Main Experimental Results}
\label{sec:main_results}

\paragraph{GOAT-Bench.}
Table~\ref{tab:exp1-main} presents the results on the full GOAT-Bench Val-Unseen split. STEGNav achieves an SR of $66.3\%$ and an SPL of $39.7$, obtaining the highest SR and the highest directly comparable SPL among all evaluated methods. Compared with MSGNav, the strongest competing method in terms of SR, STEGNav improves SR by $3.9\%$. Since MSGNav does not report SPL under the evaluation protocol adopted in this work, we compare it only in terms of SR and exclude it from the SPL comparison. Compared with EvoMemNav, the strongest baseline reporting both metrics under the same evaluation protocol, STEGNav improves SR/SPL by $6.7\%$/$0.8$. These results demonstrate that STEGNav improves the reliability of multimodal lifelong navigation while maintaining competitive path efficiency. The visualization trajectories are presented in the appendix.

\paragraph{HM3D ObjectNav.}
Table~\ref{tab:exp2-objnav} reports the results on HM3Dv1 and HM3Dv2. STEGNav achieves SR/SPL scores of $64.0\%/29.8$ on HM3Dv1 and $69.4\%/28.2$ on HM3Dv2. STEGNav obtains the highest SR on both benchmarks, outperforming the previous best results by $3.8\%$ and $5.6\%$. These consistent improvements demonstrate that the proposed spatio-temporal representation generalizes from multimodal lifelong navigation to category-level ObjectNav and supports more reliable target discovery. However, STEGNav does not achieve the best SPL on HM3D, indicating that its higher success rate does not yet fully translate into shorter navigation paths under category-level navigation settings.

\begin{table*}[t!]
\centering
\caption{Component ablation results on the GOAT-Bench Val-Unseen subset. QIG denotes query-conditioned instance grounding, OFG denotes occupancy-aware frontier grounding, and DWM denotes trajectory-aware dual-window memory. The best result in each column is highlighted in \textbf{bold}.}
\label{tab:component_ablation}
\resizebox{\textwidth}{!}{%
\begin{tabular}{cccc cc cc cc cc}
\toprule
\multicolumn{4}{c}{\textbf{Module}} &
\multicolumn{2}{c}{\textbf{Object}} &
\multicolumn{2}{c}{\textbf{Language}} &
\multicolumn{2}{c}{\textbf{Image}} &
\multicolumn{2}{c}{\textbf{Overall}} \\
\cmidrule(lr){1-4}
\cmidrule(lr){5-6}
\cmidrule(lr){7-8}
\cmidrule(lr){9-10}
\cmidrule(lr){11-12}
3D Scene Graph & QIG & OFG & DWM &
SR & SPL & SR & SPL & SR & SPL & SR & SPL \\
\midrule
\xmark & \xmark & \xmark & \xmark &
50.5 & 24.0 &41.8 & 22.5 &44.3 & 27.0 &45.7 & 24.5 \\
\cmark & \xmark & \xmark & \xmark &
61.6 & 30.5 &51.6 & 32.0 &54.5 & 38.0 &55.8 & 33.4 \\
\cmark & \cmark & \xmark & \xmark &
64.1 & 32.0 &55.6 & 35.3 &58.0 & 41.7 &59.4 & 36.2 \\
\cmark & \cmark & \cmark & \xmark &
70.7 & 37.8 &61.5 & 38.6 &68.2 & 47.5 &66.9 & 41.1 \\
\midrule
\rowcolor{blue!8}
\cmark & \cmark & \cmark & \cmark &
\textbf{74.8} & \textbf{40.8} &\textbf{63.7} & \textbf{38.7} &
\textbf{73.9} & \textbf{49.1} &\textbf{70.9} & \textbf{42.7} \\
\bottomrule
\end{tabular}}
\end{table*}

\subsection{Ablation Study}
\label{sec:ablation}

We conduct cumulative ablation experiments on a fixed GOAT-Bench subset, with all variants using the same perception modules, VLM backbone, occupancy map, low-level planner, and stopping mechanism. As shown in Table~\ref{tab:component_ablation}, introducing the 3D scene graph improves the overall SR/SPL from $45.7\%/24.5$ to $55.8\%/33.4$, validating the effectiveness of semantic representations for long-horizon navigation.

Building upon the scene graph, QIG further improves the overall SR/SPL to $59.4\%/36.2$, with more pronounced gains on language and image queries, demonstrating its ability to distinguish different physical instances within
the same semantic category. This improvement is particularly important for instance-specific queries, where category-level matching may otherwise direct the agent toward a query-irrelevant object. Adding OFG further improves the
overall SR/SPL to $66.9\%/41.1$, yielding the largest incremental gain among the proposed components. In particular, the SR/SPL for image queries increases by $10.2\%/5.8$. By explicitly associating semantic targets
with reachable frontiers and their exploration gains, OFG enables the agent to balance target relevance against exploration utility and avoid repeatedly visiting low-information regions.

Finally, DWM achieves the best overall SR/SPL of $70.9\%/42.7$. It improves the SR for image queries from $68.2\%$ to $73.9\%$, indicating that recent navigation trajectories and verified historical events provide useful evidence for instance-specific navigation. Overall, the complete model outperforms the standard 3D scene-graph variant by $15.1\%$ and $9.3$ in SR and SPL, respectively, confirming the complementary effects of QIG, OFG, and DWM.

\begin{figure}[t]
  \centering
  \includegraphics[width=\linewidth]{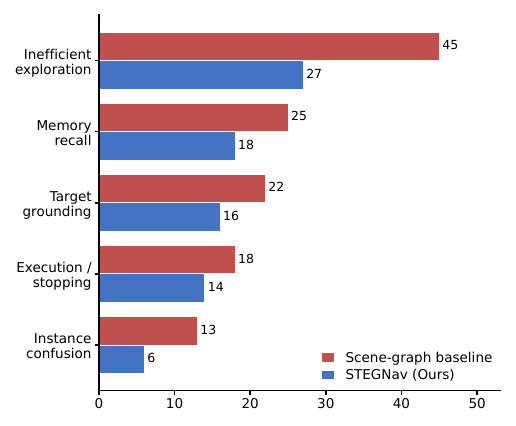}
  \caption{Failure counts of the 3D scene-graph baseline and STEGNav on the GOAT-Bench Val-Unseen subset (278).}
  \label{fig:error}
\end{figure}

\subsection{Error Analysis}
\label{sec:error_analysis}

We further categorize each failed subtask on the GOAT-Bench subset according to its primary cause, including inefficient exploration, memory recall failure, target grounding failure, execution or stopping failure, and same-category instance confusion. The primary cause is determined from the selected high-level target, recalled historical events, executed trajectory, and terminal state of each failed subtask. As shown in Fig.~\ref{fig:error},
STEGNav reduces the total number of failures from $123$ to $81$, corresponding to a $34.1\%$ reduction over the controlled scene-graph baseline. This distribution provides complementary evidence to the aggregate SR and SPL results by revealing which stages of the navigation process benefit most from the proposed representation.

Inefficient exploration exhibits the largest absolute reduction, decreasing from $45$ to $27$ failure cases. This error category is reduced by $40.0\%$, indicating that OFG and short-term trajectory memory help the agent avoid
repeatedly visiting explored regions and low-utility frontiers. Instance confusion exhibits the largest relative reduction, decreasing from $13$ to $6$ cases, corresponding to a reduction of $53.8\%$. This result validates
the effectiveness of query-conditioned instance grounding in distinguishing same-category objects. Memory recall and target grounding failures decrease from $25$ to $18$ and from $22$ to $16$, respectively, reflecting the benefits of verified event reuse and query-conditioned grounding. Execution and stopping failures decrease only from $18$ to $14$, showing a relatively limited improvement and indicating that low-level execution issues, which are not directly addressed by the proposed graph representation, remain one of the primary sources of residual errors. Overall, the error pattern shows that STEGNav primarily improves high-level grounding, exploration, and memory
reuse, while leaving additional room for improving low-level control and termination decisions.

\section{Conclusion}
\label{sec:conclusion}
We presented STEGNav, a training-free framework for multimodal lifelong navigation that extends scene graphs along spatial and temporal axes. The spatial axis constructs a query-conditioned semantic--navigational graph representing target instances, occupancy-aware frontiers, reachability, and exploration utility. The temporal axis introduces dual-window memory to preserve decision--trajectory events and verified navigation outcomes from previous subtasks. By combining these two axes, STEGNav transforms the scene graph from a static repository of observations into an event-driven environmental representation, thereby supporting instance-aware target grounding, efficient exploration, and cross-subtask experience reuse. Experiments on GOAT-Bench demonstrate that STEGNav improves both navigation success and path efficiency, while results on HM3Dv1 and HM3Dv2 further validate its generalization to category-level ObjectNav.

Future work will investigate tighter integration between high-level event-graph reasoning and low-level navigation control to further improve navigation efficiency.

\bibliography{aaai2027}

@inproceedings{habitat,
  author = {Savva, Manolis and Kadian, Abhishek and Maksymets, Oleksandr and Zhao, Yili and Wijmans, Erik and Jain, Bhavana and Straub, Julian and Liu, Jia and Koltun, Vladlen and Malik, Jitendra and others},
  title = {{Habitat}: A Platform for Embodied {AI} Research},
  booktitle = {Proceedings of the IEEE/CVF International Conference on Computer Vision},
  pages = {9339--9347},
  year = {2019}
}

@article{objectnav,
  author = {Batra, Dhruv and Gokaslan, Aaron and Kembhavi, Aniruddha and Maksymets, Oleksandr and Mottaghi, Roozbeh and Savva, Manolis and Toshev, Alexander and Wijmans, Erik},
  title = {{ObjectNav} Revisited: On Evaluation of Embodied Agents Navigating to Objects},
  journal = {arXiv preprint arXiv:2006.13171},
  year = {2020}
}

@inproceedings{vlfm,
  author = {Yokoyama, Naoki and Ha, Sehoon and Batra, Dhruv and Wang, Jiuguang and Bucher, Bernadette},
  title = {{VLFM}: Vision-Language Frontier Maps for Zero-Shot Semantic Navigation},
  booktitle = {Proceedings of the IEEE International Conference on Robotics and Automation},
  pages = {42--48},
  year = {2024}
}

@inproceedings{zson,
  author = {Majumdar, Arjun and Aggarwal, Gunjan and Devnani, Bhavika and Hoffman, Judy and Batra, Dhruv},
  title = {{ZSON}: Zero-Shot Object-Goal Navigation Using Multimodal Goal Embeddings},
  booktitle = {Advances in Neural Information Processing Systems},
  pages = {32340--32352},
  year = {2022}
}

@inproceedings{goat,
  author = {Khanna, Mukul and Ramrakhya, Ram and Chhablani, Gunjan and Yenamandra, Sriram and Gervet, Theophile and Chang, Matthew and Kira, Zsolt and Chaplot, Devendra Singh and Batra, Dhruv and Mottaghi, Roozbeh},
  title = {{GOAT-Bench}: A Benchmark for Multi-Modal Lifelong Navigation},
  booktitle = {Proceedings of the IEEE/CVF Conference on Computer Vision and Pattern Recognition},
  pages = {16373--16383},
  year = {2024}
}

@inproceedings{onemap,
  author = {Busch, Finn Lukas and Homberger, Timon and Ortega{-}Peimbert, Jes{\'u}s and Yang, Quantao and Andersson, Olov},
  title = {One Map to Find Them All: Real-Time Open-Vocabulary Mapping for Zero-Shot Multi-Object Navigation},
  booktitle = {Proceedings of the IEEE International Conference on Robotics and Automation},
  pages = {14835--14842},
  year = {2025}
}

@inproceedings{multion,
  author = {Wani, Saim and Patel, Shivansh and Jain, Unnat and Chang, Angel and Savva, Manolis},
  title = {{MultiON}: Benchmarking Semantic Map Memory Using Multi-Object Navigation},
  booktitle = {Advances in Neural Information Processing Systems},
  pages = {9700--9712},
  year = {2020}
}

@article{ssmg,
  author = {Niu, Haochen and Zhang, Lantao and Ji, Xingwu and Ying, Rendong and Liu, Peilin and Wen, Fei},
  title = {{SSMG-Nav}: Enhancing Lifelong Object Navigation with Semantic Skeleton Memory Graph},
  journal = {arXiv preprint arXiv:2603.01813},
  year = {2026}
}

@article{evomem,
  author = {Ge, Zuhao and Jia, Xiaosong and Wu, Chao and Zhou, Yuchen and Wu, Zuxuan and Jiang, Yu{-}Gang},
  title = {{EvoMemNav}: Efficient Self-Evolving Fine-Grained Memory for Zero-Shot Embodied Navigation},
  journal = {arXiv preprint arXiv:2606.03509},
  year = {2026}
}

@inproceedings{astranav,
  author = {Hu, Junjun and Xue, Xinda and Ren, Botao and Luo, Minghua and Chen, Jintao and Bai, Haochen and You, Liangliang and Xu, Mu},
  title = {{AstraNav-Memory}: Contexts Compression for Long Memory},
  booktitle = {Proceedings of the IEEE/CVF Conference on Computer Vision and Pattern Recognition},
  pages = {8097--8109},
  year = {2026}
}

@inproceedings{conceptgraphs,
  author = {Gu, Qiao and Kuwajerwala, Ali and Morin, Sacha and Jatavallabhula, Krishna Murthy and Sen, Bipasha and Agarwal, Aditya and Rivera, Corban and Paul, William and Ellis, Kirsty and Chellappa, Rama and others},
  title = {{ConceptGraphs}: Open-Vocabulary {3D} Scene Graphs for Perception and Planning},
  booktitle = {Proceedings of the IEEE International Conference on Robotics and Automation},
  pages = {5021--5028},
  year = {2024}
}

@inproceedings{scenegraphfusion,
  author = {Wu, Shun{-}Cheng and Wald, Johanna and Tateno, Keisuke and Navab, Nassir and Tombari, Federico},
  title = {{SceneGraphFusion}: Incremental {3D} Scene Graph Prediction from {RGB-D} Sequences},
  booktitle = {Proceedings of the IEEE/CVF Conference on Computer Vision and Pattern Recognition},
  pages = {7515--7525},
  year = {2021}
}

@inproceedings{sgnav,
  author = {Yin, Hang and Xu, Xiuwei and Wu, Zhenyu and Zhou, Jie and Lu, Jiwen},
  title = {{SG-Nav}: Online {3D} Scene Graph Prompting for {LLM}-Based Zero-Shot Object Navigation},
  booktitle = {Advances in Neural Information Processing Systems},
  pages = {5285--5307},
  year = {2024}
}

@inproceedings{mtu3d,
  author = {Zhu, Ziyu and Wang, Xilin and Li, Yixuan and Zhang, Zhuofan and Ma, Xiaojian and Chen, Yixin and Jia, Baoxiong and Liang, Wei and Yu, Qian and Deng, Zhidong and others},
  title = {Move to Understand a {3D} Scene: Bridging Visual Grounding and Exploration for Efficient and Versatile Embodied Navigation},
  booktitle = {Proceedings of the IEEE/CVF International Conference on Computer Vision},
  pages = {8120--8132},
  year = {2025}
}

@inproceedings{msgnav,
  author = {Huang, Xun and Zhao, Shijia and Wang, Yunxiang and Lu, Xin and Zhang, Wanfa and Qu, Rongsheng and Li, Weixin and Wang, Yunhong and Wen, Chenglu},
  title = {{MSGNav}: Unleashing the Power of Multi-modal {3D} Scene Graph for Zero-Shot Embodied Navigation},
  booktitle = {Proceedings of the IEEE/CVF Conference on Computer Vision and Pattern Recognition},
  pages = {37154--37163},
  year = {2026}
}

@inproceedings{ddppo,
  author = {Wijmans, Erik and Kadian, Abhishek and Morcos, Ari and Lee, Stefan and Essa, Irfan and Parikh, Devi and Savva, Manolis and Batra, Dhruv},
  title = {{DD-PPO}: Learning Near-Perfect {PointGoal} Navigators from 2.5 Billion Frames},
  booktitle = {Proceedings of the 8th International Conference on Learning Representations},
  year = {2020}
}

@inproceedings{aux,
  author = {Ye, Joel and Batra, Dhruv and Das, Abhishek and Wijmans, Erik},
  title = {Auxiliary Tasks and Exploration Enable {ObjectGoal} Navigation},
  booktitle = {Proceedings of the IEEE/CVF International Conference on Computer Vision},
  pages = {16117--16126},
  year = {2021}
}

@inproceedings{habitatweb,
  author = {Ramrakhya, Ram and Undersander, Eric and Batra, Dhruv and Das, Abhishek},
  title = {{Habitat-Web}: Learning Embodied Object-Search Strategies From Human Demonstrations at Scale},
  booktitle = {Proceedings of the IEEE/CVF Conference on Computer Vision and Pattern Recognition},
  pages = {5173--5183},
  year = {2022}
}

@inproceedings{semexp,
  author = {Chaplot, Devendra Singh and Gandhi, Dhiraj and Gupta, Abhinav and Salakhutdinov, Ruslan},
  title = {Object Goal Navigation Using Goal-Oriented Semantic Exploration},
  booktitle = {Advances in Neural Information Processing Systems},
  pages = {4247--4258},
  year = {2020}
}

@inproceedings{clip,
  author = {Radford, Alec and Kim, Jong Wook and Hallacy, Chris and Ramesh, Aditya and Goh, Gabriel and Agarwal, Sandhini and Sastry, Girish and Askell, Amanda and Mishkin, Pamela and Clark, Jack and others},
  title = {Learning Transferable Visual Models From Natural Language Supervision},
  booktitle = {Proceedings of the 38th International Conference on Machine Learning},
  pages = {8748--8763},
  year = {2021}
}

@inproceedings{yoloworld,
  author = {Cheng, Tianheng and Song, Lin and Ge, Yixiao and Liu, Wenyu and Wang, Xinggang and Shan, Ying},
  title = {{YOLO-World}: Real-Time Open-Vocabulary Object Detection},
  booktitle = {Proceedings of the IEEE/CVF Conference on Computer Vision and Pattern Recognition},
  pages = {16901--16911},
  year = {2024}
}

@inproceedings{cow,
  author = {Gadre, Samir Yitzhak and Wortsman, Mitchell and Ilharco, Gabriel and Schmidt, Ludwig and Song, Shuran},
  title = {{CoWs} on Pasture: Baselines and Benchmarks for Language-Driven Zero-Shot Object Navigation},
  booktitle = {Proceedings of the IEEE/CVF Conference on Computer Vision and Pattern Recognition},
  pages = {23171--23181},
  year = {2023}
}

@inproceedings{hm3dsem,
  author = {Yadav, Karmesh and Ramrakhya, Ram and Ramakrishnan, Santhosh Kumar and Gervet, Theo and Turner, John and Gokaslan, Aaron and Maestre, Noah and Chang, Angel Xuan and Batra, Dhruv and Savva, Manolis and others},
  title = {{Habitat-Matterport} {3D} Semantics Dataset},
  booktitle = {Proceedings of the IEEE/CVF Conference on Computer Vision and Pattern Recognition},
  pages = {4927--4936},
  year = {2023}
}

@inproceedings{hm3dovon,
  author = {Yokoyama, Naoki and Ramrakhya, Ram and Das, Abhishek and Batra, Dhruv and Ha, Sehoon},
  title = {{HM3D-OVON}: A Dataset and Benchmark for Open-Vocabulary Object Goal Navigation},
  booktitle = {Proceedings of the IEEE/RSJ International Conference on Intelligent Robots and Systems},
  pages = {5543--5550},
  year = {2024}
}

@article{apexnav,
  author = {Zhang, Mingjie and Du, Yuheng and Wu, Chengkai and Zhou, Jinni and Qi, Zhenchao and Ma, Jun and Zhou, Boyu},
  title = {{ApexNAV}: An Adaptive Exploration Strategy for Zero-Shot Object Navigation With Target-Centric Semantic Fusion},
  journal = {IEEE Robotics and Automation Letters},
  volume = {10},
  number = {11},
  pages = {11530--11537},
  year = {2025}
}

@inproceedings{imagegoal,
  author = {Krantz, Jacob and Gervet, Theophile and Yadav, Karmesh and Wang, Austin and Paxton, Chris and Mottaghi, Roozbeh and Batra, Dhruv and Malik, Jitendra and Lee, Stefan and Chaplot, Devendra Singh},
  title = {Navigating to Objects Specified by Images},
  booktitle = {Proceedings of the IEEE/CVF International Conference on Computer Vision},
  pages = {10916--10925},
  year = {2023}
}

@inproceedings{ieve,
  author = {Lei, Xiaohan and Wang, Min and Zhou, Wengang and Li, Li and Li, Houqiang},
  title = {Instance-Aware Exploration-Verification-Exploitation for Instance {ImageGoal} Navigation},
  booktitle = {Proceedings of the IEEE/CVF Conference on Computer Vision and Pattern Recognition},
  pages = {16329--16339},
  year = {2024}
}

@article{consistnav,
  author = {Wang, Haosen and Li, Zhenyang and Zhang, Yinqiang and He, Zongqi and Jiang, Lutao and Li, Kai and Zhao, Yizhou and Fan, Liaoyuan and Hou, Wenjian and Liang, Tingbang and others},
  title = {{ConsistNav}: Closing the Action Consistency Gap in Zero-Shot Object Navigation with Semantic Executive Control},
  journal = {arXiv preprint arXiv:2605.09869},
  year = {2026}
}

@article{aerrnav,
  author = {Huang, Jingzhi and Huang, Junkai and Yang, Haoyang and Li, Haoang and Wang, Yi},
  title = {{AERR-Nav}: Adaptive Exploration-Recovery-Reminiscing Strategy for Zero-Shot Object Navigation},
  journal = {arXiv preprint arXiv:2603.17712},
  year = {2026}
}

@inproceedings{vlmnav,
  author = {Goetting, Dylan and Singh, Himanshu Gaurav and Loquercio, Antonio},
  title = {End-to-End Navigation with {Vision-Language Models}: Transforming Spatial Reasoning into Question-Answering},
  booktitle = {Proceedings of the International Conference on Neuro-symbolic Systems},
  pages = {22--35},
  year = {2025}
}

@inproceedings{trajrag,
  author = {Wang, Yiyao and Zhang, Sixian and Zhang, Keming and Song, Xinhang and Du, Songjie and Jiang, Shuqiang},
  title = {{TrajRAG}: Retrieving Geometric-Semantic Experience for Zero-Shot Object Navigation},
  booktitle = {Proceedings of the IEEE/CVF Conference on Computer Vision and Pattern Recognition},
  pages = {15166--15176},
  year = {2026}
}

@inproceedings{glmap,
  author = {Zhang, Sixian and Wang, Yiyao and Song, Xinhang and Zhang, Keming and Xu, Zijian and Jiang, Shuqiang},
  title = {Multi-Scale Gaussian-Language Map for Zero-shot Embodied Navigation and Reasoning},
  booktitle = {Proceedings of the IEEE/CVF Conference on Computer Vision and Pattern Recognition},
  pages = {37086--37097},
  year = {2026}
}

@inproceedings{tango,
  author = {Ziliotto, Filippo and Campari, Tommaso and Serafini, Luciano and Ballan, Lamberto},
  title = {{TANGO}: Training-Free Embodied {AI} Agents for Open-World Tasks},
  booktitle = {Proceedings of the IEEE/CVF Conference on Computer Vision and Pattern Recognition},
  pages = {24603--24613},
  year = {2025}
}

@article{dynavlm,
  author = {Ji, Zihe and Lin, Huangxuan and Gao, Yue},
  title = {{DyNaVLM}: Zero-Shot Vision-Language Navigation System with Dynamic Viewpoints and Self-Refining Graph Memory},
  journal = {arXiv preprint arXiv:2506.15096},
  year = {2025}
}

@inproceedings{dsg,
  author = {Rosinol, Antoni and Gupta, Arjun and Abate, Marcus and Shi, Jingnan and Carlone, Luca},
  title = {{3D} Dynamic Scene Graphs: Actionable Spatial Perception with Places, Objects, and Humans},
  booktitle = {Proceedings of Robotics: Science and Systems},
  year = {2020}
}

@inproceedings{unigoal,
  author = {Yin, Hang and Xu, Xiuwei and Zhao, Linqing and Wang, Ziwei and Zhou, Jie and Lu, Jiwen},
  title = {{UniGoal}: Towards Universal Zero-Shot Goal-Oriented Navigation},
  booktitle = {Proceedings of the IEEE/CVF Conference on Computer Vision and Pattern Recognition},
  pages = {19057--19066},
  year = {2025}
}

@inproceedings{saynav,
  author = {Rajvanshi, Abhinav and Sikka, Karan and Lin, Xiao and Lee, Bhoram and Chiu, Han-Pang and Velasquez, Alvaro},
  title = {{SayNav}: Grounding Large Language Models for Dynamic Planning to Navigation in New Environments},
  booktitle = {Proceedings of the 34th International Conference on Automated Planning and Scheduling},
  pages = {464--474},
  year = {2024}
}

@inproceedings{hsan,
  author = {Fang, Xiang and Fang, Wanlong and Wang, Changshuo},
  title = {Hierarchical Semantic-Augmented Navigation: Optimal Transport and Graph-Driven Reasoning for Vision-Language Navigation},
  booktitle = {Advances in Neural Information Processing Systems},
  year = {2025}
}

@inproceedings{3dmem,
  author = {Yang, Yuncong and Yang, Han and Zhou, Jiachen and Chen, Peihao and Zhang, Hongxin and Du, Yilun and Gan, Chuang},
  title = {{3D-Mem}: {3D} Scene Memory for Embodied Exploration and Reasoning},
  booktitle = {Proceedings of the IEEE/CVF Conference on Computer Vision and Pattern Recognition},
  pages = {17294--17303},
  year = {2025}
}

@inproceedings{l3mvn,
  author = {Yu, Bangguo and Kasaei, Hamidreza and Cao, Ming},
  title = {{L3MVN}: Leveraging Large Language Models for Visual Target Navigation},
  booktitle = {Proceedings of the IEEE/RSJ International Conference on Intelligent Robots and Systems},
  pages = {3554--3560},
  year = {2023}
}

@inproceedings{pixnav,
  author = {Cai, Wenzhe and Huang, Siyuan and Cheng, Guangran and Long, Yuxing and Gao, Peng and Sun, Changyin and Dong, Hao},
  title = {Bridging Zero-Shot Object Navigation and Foundation Models through Pixel-Guided Navigation Skill},
  booktitle = {Proceedings of the IEEE International Conference on Robotics and Automation},
  pages = {5228--5234},
  year = {2024}
}

@article{spnet,
  author = {Zhao, Qianfan and Zhang, Lu and He, Bin and Liu, Zhiyong},
  title = {Semantic Policy Network for Zero-Shot Object Goal Visual Navigation},
  journal = {IEEE Robotics and Automation Letters},
  volume = {8},
  number = {11},
  pages = {7655--7662},
  year = {2023}
}

@inproceedings{sgm,
  author = {Zhang, Sixian and Yu, Xinyao and Song, Xinhang and Wang, Xiaohan and Jiang, Shuqiang},
  title = {Imagine Before Go: Self-Supervised Generative Map for Object Goal Navigation},
  booktitle = {Proceedings of the IEEE/CVF Conference on Computer Vision and Pattern Recognition},
  pages = {16414--16425},
  year = {2024}
}

@inproceedings{compassnav,
  author = {Li, LinFeng and Zhao, Jian and Xie, Yuan and Tan, Xin and Li, Xuelong},
  title = {{CompassNav}: Steering from Path Imitation to Decision Understanding in Navigation},
  booktitle = {Proceedings of the 14th International Conference on Learning Representations},
  year = {2026}
}

@inproceedings{sam,
  author = {Kirillov, Alexander and Mintun, Eric and Ravi, Nikhila and Mao, Hanzi and Rolland, Chloe and Gustafson, Laura and Xiao, Tete and Whitehead, Spencer and Berg, Alexander C. and Lo, Wan{-}Yen and others},
  title = {Segment Anything},
  booktitle = {Proceedings of the IEEE/CVF International Conference on Computer Vision},
  pages = {4015--4026},
  year = {2023}
}

% Include the supplementary source in the same LaTeX document.
\clearpage
\raggedbottom
\section*{Supplementary Material}

\section{Implementation Details}
\label{app:supplementary}

This appendix complements the main paper with the implementation choices required
to instantiate STEGNav. We focus on the method-level procedure, the principal
hyperparameters, prompts used for VLM reasoning and case study.

\subsection{Overall Navigation Procedure}
\label{app:overall_procedure}

Algorithm~\ref{alg:app_stegnav} summarizes the complete online procedure. At each
high-level step, the agent first updates the semantic scene graph and occupancy map
from a set of egocentric RGB-D observations. The spatial axis then grounds the
current query to candidate instances and augments the graph with occupancy-aware
frontier nodes. The temporal axis reconnects recent decision--trajectory events and
verified historical anchors to the current graph. A single VLM decision subsequently
selects either an object instance, an exploration frontier, or an object visible in
an earlier observation. The executed trajectory and its outcome are finally written
back to the temporal representation.

\begin{algorithm}[ht]
\caption{STEGNav Inference}
\label{alg:app_stegnav}
\small
\begin{algorithmic}[1]
\Require Queries $q_1,\ldots,q_K$, graph $\mathcal{G}$, occupancy map
$\mathbf{O}^{\mathrm{occ}}$, long-term memory $\mathcal{M}^{\mathrm{lt}}$
\For{$k=1,\ldots,K$}
    \State $\mathcal{M}^{\mathrm{st}}\gets\varnothing$
    \For{$t=1,\ldots,B$}
        \State $(\mathcal{G}_t,\mathbf{O}^{\mathrm{occ}}_t)
        \gets\textsc{MapUpdate}(\textsc{Observe}())$
        \State $(\mathcal{C}_{k,t},\mathcal{E}^{\mathrm{qry}}_{k,t})
        \gets\textsc{InstanceGround}(q_k,\mathcal{G}_t)$
        \State $(\mathcal{V}^{\mathrm{fr}}_t,\mathcal{E}^{\mathrm{nav}}_t)
        \gets\textsc{FrontierGround}(\mathbf{O}^{\mathrm{occ}}_t,\mathcal{G}_t)$
        \State $\mathcal{Z}_{k,t}\gets\textsc{TemporalLink}
        (\mathcal{M}^{\mathrm{st}},\mathcal{M}^{\mathrm{lt}},\mathcal{G}_t)$
        \State $\widetilde{\mathcal{H}}_{k,t}\gets\textsc{ComposeEventGraph}
        (\mathcal{C}_{k,t},\mathcal{V}^{\mathrm{fr}}_t,\mathcal{Z}_{k,t})$
        \State $y_{k,t}\gets\pi_{\mathrm{VLM}}
        (q_k,\operatorname{Ser}(\widetilde{\mathcal{H}}_{k,t}))$
        \State $(\tau_{k,t},o_{k,t})\gets\textsc{Navigate}(y_{k,t})$
        \State $\mathcal{M}^{\mathrm{st}}\gets\textsc{Append}
        (\mathcal{M}^{\mathrm{st}},y_{k,t},\tau_{k,t},o_{k,t})$
        \If{$\textsc{Verify}(q_k,o_{k,t})=\mathrm{positive}$}
            \State $\mathcal{M}^{\mathrm{lt}}\gets\textsc{StoreAnchor}
            (\mathcal{M}^{\mathrm{lt}},q_k,y_{k,t},o_{k,t})$
            \State \textbf{break}
        \EndIf
    \EndFor
\EndFor
\end{algorithmic}
\end{algorithm}

The step budget adapts to scene scale:
\begin{equation}
    B=\max\!\left(50,\left\lfloor 2\sqrt{A}\right\rfloor\right),
    \label{eq:app_step_budget}
\end{equation}
where $A$ is the horizontal area of the scene bounding box in square metres. An
object commitment may persist across multiple motion segments, whereas a frontier
commitment is reconsidered after the next map update.

\subsection{Implementation Settings}
\label{app:settings}

\subsubsection{Perception and Online Mapping}

Each high-level step uses seven RGB-D observations: the current heading and six
additional headings separated by $40^\circ$. The simulator resolution is
$1280\times1280$, with a $120^\circ$ horizontal field of view, a camera height of
$1.5$ m, and a camera tilt of $-30^\circ$. Images passed to the VLM are resized to
$512\times512$.

We use YOLOv8x-World for open-vocabulary detection, SAM-L for instance masks, and
CLIP ViT-H-14-quickgelu with \texttt{dfn5b} weights for appearance features. The
detector vocabulary is constructed from the semantic categories available in each
HM3D scene. The scene graph associates multi-view detections into persistent 3D
object nodes and connects nearby nodes within $3.5$ m. The occupancy representation
uses a TSDF grid with $0.1$ m resolution. No simulator semantic label or ground-truth
goal position is supplied to the navigation policy.

\subsubsection{Query-Conditioned Instance Grounding}
\label{app:qig}

Instance grounding is applied only to image and language-description queries. A
category-level VLM prefilter first retains at most 20 relevant scene nodes. CLIP then
pre-ranks these nodes and limits the grounding prompt to at most 10 candidate
instances. The grounding VLM predicts a relevance score $r_k^t(i)\in[0,1]$ and a
query-conditioned description $d_k^t(i)$ for each candidate. We retain the five
highest-scoring instances and represent them as query--instance edges. If VLM
grounding is unavailable, CLIP similarity provides the relevance fallback.

After the navigation VLM selects an object node, a same-class CLIP re-ranking stage
is used for image and description queries. The selected node is changed only when
another candidate improves the similarity by at least $0.05$. Category-level object
queries bypass instance pruning and re-ranking because any instance of the requested
category is valid.

\subsubsection{Occupancy-Aware Frontier Grounding}
\label{app:ofg}

Frontier cells are extracted at the boundary between agent-connected free space and
unexplored space and clustered into frontier regions. Clusters spanning more than
$150^\circ$ around the agent are subdivided to avoid representing several exploration
directions as one node. The minimum frontier size is 20 grid cells for GOAT-Bench and
10 cells for HM3D.

Each frontier node contains its current centre, reachability, geodesic path cost,
exploration gain, visitation state, and a facing RGB view. Exploration gain is the
unexplored area within a $1.0$ m dilation of the frontier region. A frontier is also
connected to at most three semantic anchors within $2.5$ m on the current traversable
island. Agent-to-object edges contain the same reachability and path-cost attributes,
allowing semantic targets and exploration frontiers to be compared in a common
representation.

The serialized spatial graph is ordered as query event, agent node, object nodes,
frontier nodes, object relations, and visual evidence. Object coordinates and
frontier coordinates are rounded to two decimal places. Relation images are selected
by greedy coverage of the retained scene-graph edges.

\subsubsection{Trajectory-Aware Dual-Window Memory}
\label{app:memory}

The short-term window retains the latest $W_{\mathrm{st}}=20$ executed decisions.
Each event stores the selected node, trajectory endpoints, travelled distance,
traversed semantic regions, frontier coverage when applicable, and the execution
outcome. Repeated visits remain separate events, preserving the actual temporal
sequence. The window is cleared at the end of each subtask.

The long-term window retains at most $W_{\mathrm{lt}}=8$ verified outcomes across
subtasks in a GOAT-Bench episode. An event is admitted when
\begin{equation}
\begin{aligned}
    w_k&=\mathbf{1}\!\left[
        i_k^{\mathrm{term}}=i_k^* \land o_k=+ \land d_k\le\delta
    \right],\\
    \delta&=0.8\ \mathrm{m},
\end{aligned}
    \label{eq:app_verification}
\end{equation}
where $i_k^{\mathrm{term}}$ is the instance associated with the terminal observation,
$o_k$ is the VLM verification result, and $d_k$ is the terminal distance to the
selected node. This verification uses the agent's observations and current scene
graph rather than simulator ground truth.

Historical anchors are re-associated with live scene nodes using persistent
identifiers and CLIP appearance similarity. We accept an appearance match when
$\rho\ge0.92$, reject it when $\rho\le0.70$, and use a VLM identity verifier for
ambiguous cases. A spatial association radius of $0.6$ m is used when comparing live
same-class nodes. To avoid transferring category-level evidence to an incompatible
instance-specific task, an anchor created by an object-category query cannot validate
a later image or description query.

\subsubsection{Planning and VLM Inference}

The selected goal is projected to the nearest valid cell on the agent-connected
traversable island. GOAT-Bench uses a $1.0$ m motion horizon for both object and
frontier goals. HM3D uses $1.0$ m for object goals and $1.8$ m for frontier goals.
After reaching a non-frontier goal, the VLM verifies the target using observations
from at most the latest five high-level steps. Benchmark success is measured
separately using the standard $1.0$ m geodesic criterion.

All reasoning modules use the same VLM backbone, \texttt{gpt-5.4-mini}. Decoding uses
temperature $0.95$, a maximum of 4096 output tokens, and one completion per request.
Thus, STEGNav does not use majority voting or self-consistency sampling.

\subsection{Core Prompt Templates}
\label{app:prompts}

The following prompt boxes present the main prompt content. Angle-bracketed fields
are filled at runtime, and image markers denote multimodal image parts. Fixed
demonstrations and repeated graph rows are omitted for conciseness.

\begin{promptbox}{Prompt 1: Category-Level Scene-Graph Prefilter}
\promptrole{System Prompt}
You are an AI agent in a 3D indoor scene.

\tcblower
\promptrole{User Prompt}
Identify the scene-graph objects that are most helpful for locating the target.
Each object contains an identifier, class, room, and neighbouring object
identifiers.

\medskip
\noindent\textbf{Selection criteria:}
\begin{itemize}[leftmargin=1.3em, itemsep=2pt, topsep=3pt, parsep=0pt]
    \item semantic relevance to the target;
    \item typical co-occurrence with the target;
    \item likelihood of being in the same room; and
    \item spatial diversity among the selected nodes.
\end{itemize}

\medskip
\noindent\textbf{Task input:}

\noindent Question: \texttt{<QUESTION>} \\
\texttt{<REFERENCE\_IMAGE>}, for image queries only.

\medskip
\noindent\textbf{Scene-graph format:}

\noindent\texttt{<ID>: <CLASS>, <ROOM>, [<NEIGHBOR\_ID>, \ldots]}

\medskip
\noindent\textbf{Output format:}
Return only the selected object identifiers, one per line, in ranked order. Do
not output an identifier that is absent from the input graph.
\end{promptbox}

\begin{promptbox}{Prompt 2: Query-Conditioned Instance Grounding}
\promptrole{System Prompt}
You are grounding a navigation query onto individual object instances in an
indoor scene. Several instances of the same category may exist, and only one may
match the query. Judge each candidate independently using its appearance and
spatial context.

\tcblower
\promptrole{User Prompt}
\noindent\textbf{Query:}

\noindent Target category: \texttt{<CLASS>} \\
Language query: \texttt{<QUESTION>}, or reference image:
\texttt{<REFERENCE\_IMAGE>}.

\medskip
\noindent\textbf{Candidate format:}

\noindent\texttt{Instance <ID>: <CLASS>, room <ROOM>,} \\
\texttt{position (<X>, <Z>, <Y>), near <IDs>} \\
\texttt{<INSTANCE\_VIEW>}

\medskip
\noindent\textbf{Hard rules:}
\begin{itemize}[leftmargin=1.3em, itemsep=2pt, topsep=3pt, parsep=0pt]
    \item For an image query, assess whether the candidate view shows the same
    physical instance as the reference image.
    \item For a description query, assess whether appearance, room, and spatial
    context satisfy the description.
\end{itemize}

\medskip
\noindent\textbf{Output format:}
Output exactly one line per candidate:
\texttt{<id> <relevance> <description>}. The relevance must lie in $[0,1]$,
and the description must briefly state the supporting visual evidence. Output no
additional text.
\end{promptbox}

\begin{promptbox}{Prompt 3: Spatio-Temporal Event-Graph Navigation}
\promptrole{System Prompt}
You are an agent navigating an indoor scene. Reason over a spatio-temporal event
graph and choose the next navigation goal. Find the requested target within
\texttt{<B>} high-level steps.

\tcblower
\promptrole{User Prompt}
\noindent\textbf{Task input:}

\noindent Question: \texttt{<QUESTION>} \\
\texttt{<REFERENCE\_IMAGE>}, for image queries only.

\medskip
\noindent\textbf{Event-graph components:}
The graph contains a \textbf{Query Event} with the target and modality; an
\textbf{Agent Node} with the current position and budget; \textbf{Object Nodes}
with grounding and navigation attributes; \textbf{Frontier Nodes} with exploration
and navigation attributes; image-supported \textbf{Object--Object Relations}; and
\textbf{Recent Events} describing executed decisions, trajectories, and outcomes.

\medskip
\noindent\textbf{Decision rules:}
\begin{enumerate}[leftmargin=1.5em, itemsep=2pt, topsep=3pt, parsep=0pt]
    \item Select an object whose visual evidence matches the query, favouring high
    relevance and low path cost while verifying predicted labels against images.
    \item Otherwise, select a reachable frontier by balancing exploration gain,
    path cost, semantic context, and visitation.
    \item Select an earlier image only when it shows the target but no object node
    represents it; include the object category.
    \item Use recent events as evidence, not as hard constraints against revisits.
\end{enumerate}

\medskip
\noindent\textbf{Runtime context:}

\noindent\texttt{<SERIALIZED\_EVENT\_GRAPH>} \\
\texttt{<ORDERED\_IMAGE\_LIST>}

\medskip
\noindent\textbf{Output format:}
Output exactly one of \texttt{Object <ID>}, \texttt{Frontier <ID>}, or
\texttt{Image <ID>, <CLASS>} on the first line.
Optional brief reasoning may follow on a new line.
\end{promptbox}

\begin{promptbox}{Prompt 4: Stop Verification}
\promptrole{System Prompt}
You are an indoor navigation agent verifying whether the requested target has
been reached. Base the decision only on the query and the provided terminal
observations.

\tcblower
\promptrole{User Prompt}
\noindent Question: \texttt{<QUESTION>} \\
\texttt{<REFERENCE\_IMAGE>}, for image queries only.

\medskip
Inspect the surrounding observations from at most the latest five steps and
determine whether the agent has reached the target required by the query.

\medskip
\noindent\texttt{<TERMINAL\_OBSERVATIONS>}

\medskip
\noindent\textbf{Output format:}
Answer \texttt{Yes} or \texttt{No} on the first line. Optional brief reasoning
may follow on a new line.
\end{promptbox}

\begin{promptbox}{Prompt 5: Ambiguous Historical Identity}
\promptrole{System Prompt}
You are comparing two navigation goals from the same household episode. Both
refer to the same object category. Decide whether they describe the same physical
instance or different instances.

\medskip
\noindent\textbf{Hard rules:}
\begin{itemize}[leftmargin=1.3em, itemsep=2pt, topsep=3pt, parsep=0pt]
    \item Compare instance-specific appearance and contextual evidence.
    \item Do not infer identity solely from the shared category name.
    \item Use only information contained in the two queries and reference images.
\end{itemize}

\tcblower
\promptrole{User Prompt}
\noindent\textbf{Goal A: previously reached} \\
Description: \texttt{<OLD\_QUERY>} \\
\texttt{<OLD\_REFERENCE\_IMAGE>}, if present.

\medskip
\noindent\textbf{Goal B: current query} \\
Description: \texttt{<NEW\_QUERY>} \\
\texttt{<NEW\_REFERENCE\_IMAGE>}, if present.

\medskip
\noindent\textbf{Output format:}
Answer \texttt{Yes} or \texttt{No} on the first line. \texttt{Yes} means the two
goals refer to the same physical instance. Optional brief reasoning may follow on
a new line.
\end{promptbox}

\section{Case Study}

\begin{figure*}[t]
  \centering
  \includegraphics[width=\linewidth]{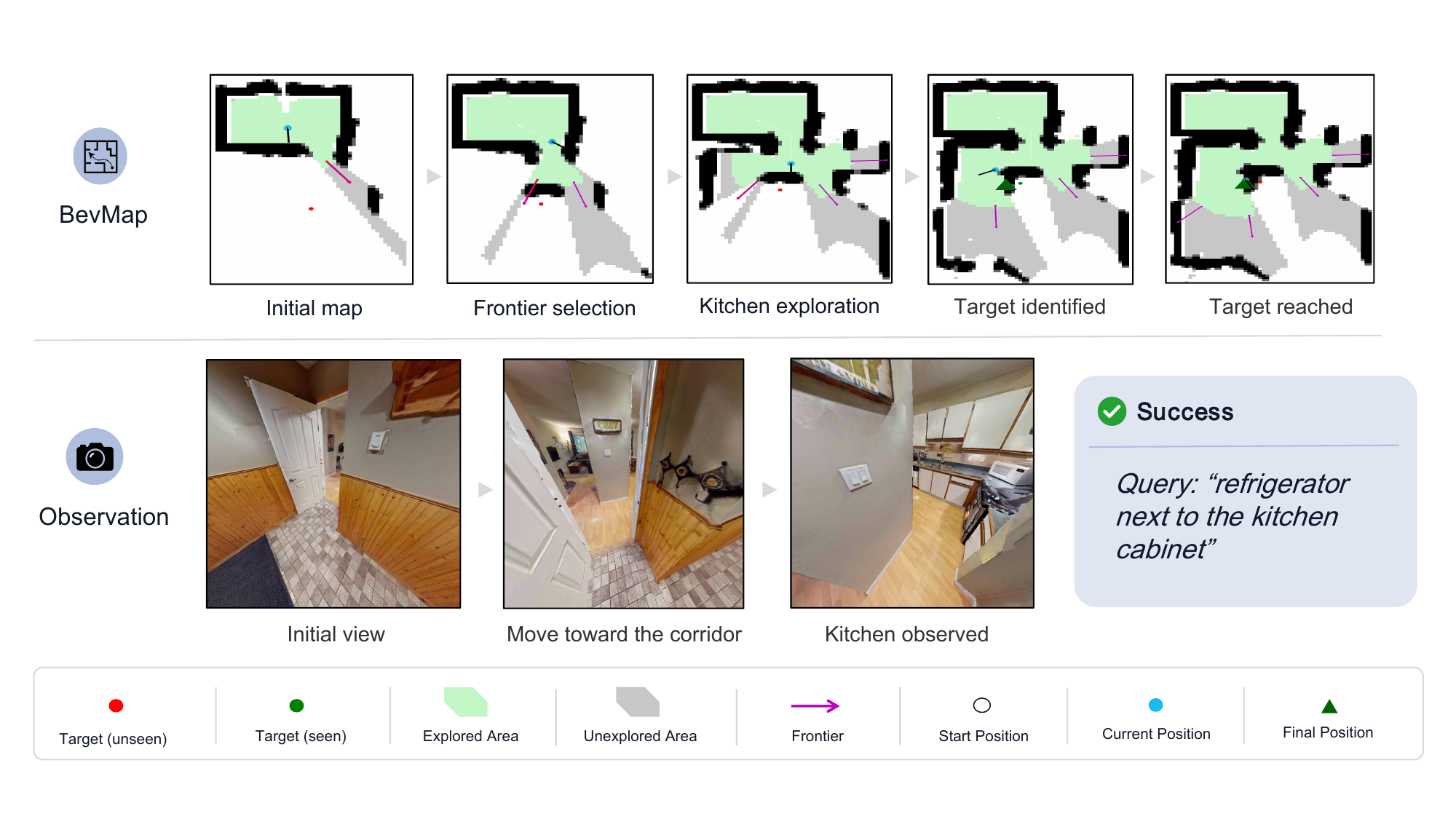}
  \caption{Successful language-guided navigation. Given the query \emph{``refrigerator next to the kitchen cabinet''}, STEGNav progressively explores informative regions and grounds the target instance using semantic and spatial cues. The trajectory achieves an SPL of 0.587 with only 4 navigation steps, reaching the target within 0.40 m.}
  \label{fig:case-refrigerator}
\end{figure*}

\begin{figure*}[t]
  \centering
  \includegraphics[width=\linewidth]{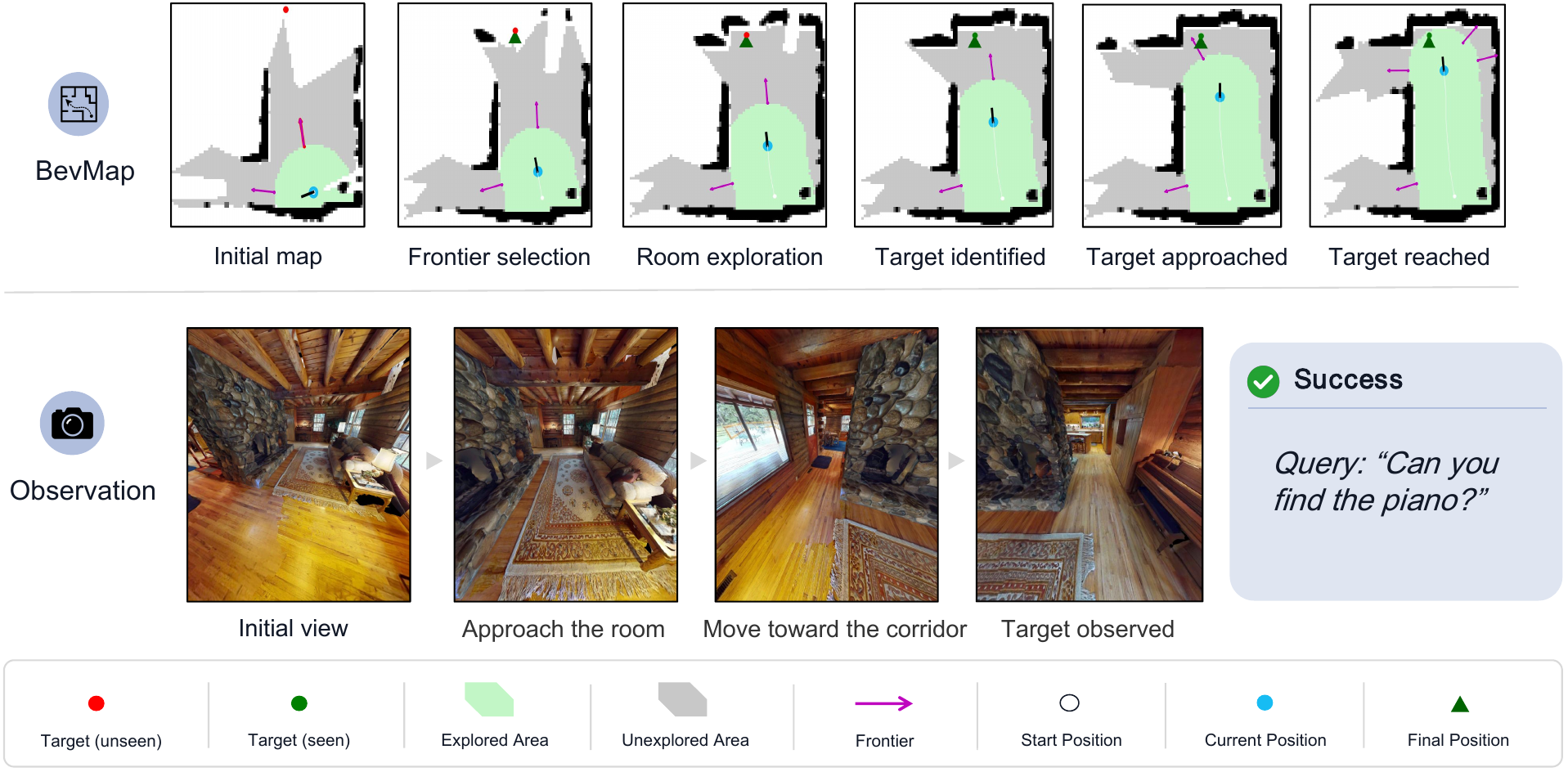}
  \caption{Successful object navigation. Given the query \emph{``Can you find the piano?''}, STEGNav selects informative frontiers, discovers the target region, and successfully reaches the target. The trajectory achieves an SPL of 0.913 with 6 navigation steps, reaching the target within 0.97 m.}
  \label{fig:case-piano}
\end{figure*}

We present two successful navigation examples in Figures~\ref{fig:case-refrigerator} and~\ref{fig:case-piano}. 
Each visualization combines the bird's-eye-view (BEV) trajectory with corresponding RGB observations, allowing inspection of how STEGNav progressively explores unseen environments, reasons over navigation targets, and executes the final approach. 
The first case focuses on language-guided instance grounding, where the agent must identify a target object described by both semantic and spatial attributes. 
The second case illustrates object-level navigation, where the agent discovers the target through frontier-based exploration in an unseen environment. 
Together, these examples demonstrate how STEGNav integrates semantic understanding and exploration reasoning to achieve reliable multimodal navigation.

\paragraph{Language-guided target grounding.}
Figure~\ref{fig:case-refrigerator} illustrates a successful navigation trajectory given the query \emph{``refrigerator next to the kitchen cabinet.''}
Starting from a partially explored environment, STEGNav first selects informative frontiers to expand the explored region. 
After reaching the kitchen area, the agent identifies the target instance by jointly reasoning over semantic and spatial information, and finally navigates to a valid target viewpoint. 
This case demonstrates that the spatial axis of STEGNav enables effective target grounding by integrating object semantics with environmental context.

\paragraph{Object navigation with exploration.}
Figure~\ref{fig:case-piano} presents a successful object navigation example for the query \emph{``Can you find the piano?''}
Without prior knowledge of the complete environment, STEGNav incrementally expands the explored area through frontier selection and identifies the target after entering the relevant room. 
The agent then approaches the target and terminates at a suitable viewpoint. 
This example highlights the ability of STEGNav to unify exploration and target-oriented navigation in unseen environments.

\end{document}